\documentclass[letterpaper, 10 pt, conference]{ieeeconf}  

\IEEEoverridecommandlockouts                              
\usepackage{amsmath} 
\usepackage{amssymb}  
\usepackage{url}

\usepackage{subcaption}
\usepackage{graphicx}
\usepackage{algorithm}
\usepackage{algorithmic}

\title{\LARGE \bf
Fusion Strategies for Learning User Embeddings with Neural Networks
}

\author{Philipp Blandfort$^{1,2}$, Tushar Karayil$^1$, Federico Raue$^1$, J{\"o}rn Hees$^1$ and Andreas Dengel$^{1,2}$ \\
email: \texttt{firstname.lastname@dfki.de} 
\thanks{$^{1}$German Research Center for Artificial Intelligence (DFKI), Kaiserslautern, Germany
       }%
\thanks{$^{2}$Technische Universit{\"a}t
        Kaiserslautern, Germany
   }%
}

\begin{document}

\maketitle
\thispagestyle{empty}
\pagestyle{empty}

\begin{abstract}

Growing amounts of online user data motivate the need for automated processing techniques.
In case of user ratings, one interesting option is to use neural networks for learning to predict ratings given an item and a user.
While training for prediction, such an approach at the same time learns to map each user to a vector, a so-called user embedding.
Such embeddings can for example be valuable for estimating user similarity.
However, there are various ways how item and user information can be combined in neural networks, and it is unclear how the way of combining affects the resulting embeddings.

In this paper, we run an experiment on movie ratings data, where we analyze the effect on embedding quality caused by several fusion strategies in neural networks.
For evaluating embedding quality, we propose a novel measure, Pair-Distance Correlation, which quantifies the condition that similar users should have similar embedding vectors.
We find that the fusion strategy affects results in terms of both prediction performance and embedding quality.
Surprisingly, we find that prediction performance not necessarily reflects embedding quality.
This suggests that if embeddings are of interest, the common tendency to select models based on their prediction ability
should be reconsidered.
\end{abstract}


\section{INTRODUCTION} \label{sec:introduction}

The past two decades have seen an exponential proliferation of user-generated content across the Internet, including social media posts, user activities and ratings.
Such user data has been used 
in a variety of ways.
Examples include the detection of users' sentiment from product reviews \cite{tang2009survey}, but
user data has also been used to train models for predicting where users will click \cite{graepel2010web} or which items they will like \cite{ricci2015recommender}.
Such detection and prediction tasks typically have direct practical motivations.
It can, however, be important as well to add an explanatory component to such detection and prediction systems.
In particular, this importance can be due to legal reasons, since the European legislation (GDPR \cite{EU-2016-679-GDPR}) now grants users the right to ask for a simple explanation of any automatic decisions that affect them.
There are several possibilities for combining analysis with detection or prediction, in order to make AI systems more understandable.

One way, for example, is to build on understandable mid-level concepts for detection, such that the trained model automatically has an explanatory quality.
This approach is adopted in aspect-based sentiment detection (e.g., \cite{liu2012survey}), which is commonly applied to user reviews to not only detect the overall sentiment towards a product but simultaneously describe which aspects of the product are responsible for the user's opinion. 
Another approach is to fit the given data with a complex model (often a neural network) and then derive explanations from the trained model by means of sophisticated analysis techniques.
This direction includes recent efforts related to heatmapping techniques that are used for explaining decisions of machine learning models (e.g., \cite{MONTAVON20181,SamITU18}).
For example, in the field of medicine, Sturm et al. \cite{sturm2016interpretable} show how to train a neural network on classifying EEG data and then use a heatmapping technique to generate explanations for the network's classification decisions.
%
Yet another case of combining prediction with analysis would be representation learning, where some concept of interest (such as the user) is mapped to a vector as part of a larger model that fits the given data. An example for such an approach is proposed by Amir et al. \cite{amir2017quantifying}. In their paper, users are embedded to a vector and then fused into a neural network for predicting textual contents (twitter texts).
In this regard,  neural networks represent an interesting choice as a model.  This is because, in the past few years,  many techniques have been proposed for analyzing them (such as the ones mentioned above). 

We see that most of the mentioned approaches involve training a neural network on user data, which combines user and item information for prediction.
Several strategies exist for such an information fusion, but so far the effect of this choice has barely been analyzed.

\begin{figure}[t]
\begin{center}
    \includegraphics[width=0.7\linewidth]{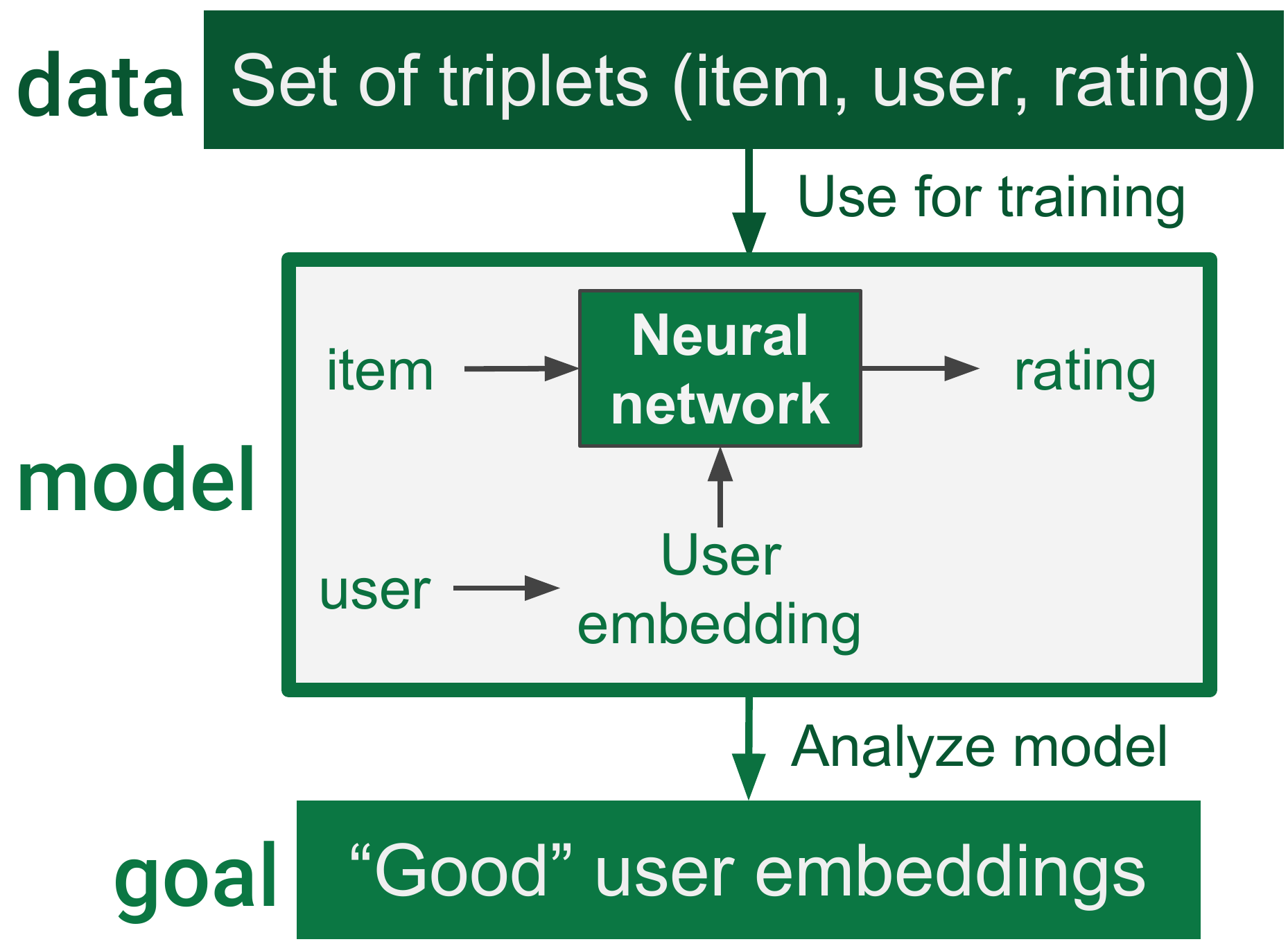}
\end{center}
\caption{Illustration of the neural-network-based approach adopted in this paper for learning user embeddings.
A neural network learns to fit the data while simultaneously embedding the users, based on item ratings from different users.
Our main questions are: How to quantitatively evaluate the learned embeddings?
And what are the effects of fusing item and user information in a particular way?} \label{fig:diagram}
\end{figure}

Hence, in this work we explore the direction of using neural networks for simultaneously fitting user data and learning vector representations for users,
and analyze the effects of different fusion variants.
As user data, we decide to use movie ratings from different users, where we represent movies as (dense) feature vectors based on tags.
Our goal is to fit this data with a neural network that takes user ID and movie features as input, while predicting the corresponding ratings and learning to embed the users into vectors.
The focus thereby lies on the representation learning component, i.e., we want to find out how to learn ``good'' embeddings.
To this end, we mainly address these questions:
Which way of combining user and movie information is suitable for such a task?
What is the effect of embedding size? And how can we define and quantify the quality of embeddings?
In this regard, the contributions of this work can be summed up as following:

\begin{enumerate}
  \item We propose a novel evaluation measure for quality of user embeddings. 
  \item We analyze the effect of various fusion strategies and embedding size on the resulting quality of learned embeddings.
\end{enumerate}

The rest of the paper is organized as follows. Section \ref{sec:related_work} summarizes the previous works that have been related to this field of research.
Section \ref{basics} formalizes the task in a detailed manner and introduces the proposed measure for embedding quality.
Section \ref{sec:methods} explains the relevant fusion strategies for neural networks. 
Section \ref{sec:experiments} describes the experiment performed.
Section \ref{conclusion} summarizes our findings and mentions some future directions.

\section{RELATED WORK}\label{sec:related_work}

\subsection{Learning user embeddings}
The main goal of this paper is to learn (meaningful) user embeddings.
%
%
In the literature, user information has been used in various ways.
For example, user information can be exploited for adding cognitive information to the model.
This was done by Yamagashi et al. \cite{yamagishi2004speaking} who defined a user embedding as context for learning different speaking styles, such as reading, joyful, and sad.
There, the user information is defined in terms of two components: phonetic and linguistic.
Additionally, user information has been used for detecting sarcasm and mental health conditions based on social media data \cite{amir2016modelling,amir2017quantifying}.
In both of these scenarios, neural networks based on \texttt{Paragraph2Vec} \cite{le2014distributed} are trained on textual contents with the goal of learning user-dependent word-usage patterns.
In this process, the model automatically learns user embeddings that are based on the relationship between users and their texts. 

Both of these papers analyze the learned user embeddings in order to obtain insights about user behavior. 
The authors, however, do not propose a formal measure for evaluating embedding quality, and the fusion strategy in both cases is simple concatenation.
We will propose a novel measure for quantitative evaluation of embedding quality in Section~\ref{sec:rank_measure}.
Also, whether concatenation is the most appropriate fusion strategy for learning embeddings is far from obvious.
Indeed, we can find several other fusion strategies for neural networks in different areas, which we shall briefly discuss now.

\subsection{Fusion strategies in neural networks}
Fusion strategies have been applied to two or more modalities for joining representations and predictions.
One possibility is to have a shared representation in the model \cite{ngiam2011multimodal}.
Furthermore, each modality is learned individually as a first layer and then both components are joined into a shared representation as a second layer.  
This can be seen as a early fusion. 
Additionally, the fusion can also be presented in the middle or at the end of the model.
For example, a common approach in Visual-Questioning Answering (VQA) is to first obtain visual and text embeddings after applying a Convolutional Neural Network (CNN) and Recurrent Neural Network (RNN), respectively \cite{Teney_2018_CVPR}.
Then, a simple Hadamard product (i.e. element-wise multiplication) is used as a fusion method in the model.
The previous two approaches are based on concatenation or multiplication operations.
Another approach is to apply tensor operations to the multimodal embeddings.
For example, the MUTAN model \cite{ben2017mutan} factorizes a multimodal tensor generated by the question and image embeddings.
Similarly, tensor products are used for information fusion in some works of distributional semantics (e.g., \cite{guevara2010regression, bamman2014distributed, hartung2017learning}).
Some of these works (\cite{guevara2010regression, hartung2017learning}) even include a systematic comparison of fusion strategies in terms of effects on embeddings of adjective-noun combinations.
However, for the case of user embeddings, no such comparative study exists. 

\section{PRELIMINARIES} \label{basics}

\subsection{Problem statement} \label{sec:problem}

We assume that we are given data of the form $\{ (x_1,u_1,R_1), \ldots, (x_m,u_m,R_m) \}$, where $x_i \in \mathbb{R}^n$ are input items (such as movies), $u_i \in \mathcal{U}$ correspond to users with $\mathcal{U}$ the space of user identifiers, and $R_i \in \mathbb{R}$ is the rating which the user $u_i$ assigned to item $x_i$.
The goal is to find a mapping $e: \mathcal{U} \rightarrow \mathbb{R}^z$ that assigns each user to a real-valued vector, which we refer to as the user's \textit{embedding}.
For example, a user with ID \texttt{user\_1} could be mapped to a 3-D vector $e(\texttt{user\_1}) = [0.2, 0.1, 0.7]^T$, where $z=3$.
We require that these embeddings are ``meaningful'' in the sense that similarity of embeddings should reflect similarity between users.
Intuitively, we want to represent the users such that it is easy for us to see how similar they are in terms of how they rate items.
(This part will be formalized as a novel measure in Section~\ref{sec:rank_measure}.)

In this paper, we analyze how this goal can be achieved by fitting the given data with a neural network that simultaneously learns embeddings for the users.
The main questions we address are: How does embedding size $z$ relate to the quality of the learned embeddings and the ability to fit the data? And, since such a neural network needs to combine input item and user information for predicting a rating, which fusion strategy is most appropriate for this task?

\subsection{Functional Data Analysis (FDA)} 

In order to foster a deeper understanding of the problem, we describe its relation to a particular branch of mathematics, namely Functional Data Analysis (FDA).
This will show how learning user embeddings can be understood as finding vector representations for functions.
This insight will then be useful later for seeing how embeddings can be evaluated quantitatively.

Mathematically, we can model the data generation process analogously to the modeling in FDA (compare, e.g., with the description in the survey of Jacques and Preda \cite{jacques2014functional}):
We assume that there is a functional random variable
$$ F: \Omega \rightarrow \{f: I \rightarrow \mathbb{R}\} \,, $$ 
i.e., $F$ is a random variable which has functions (from $I$ to $\mathbb{R}$) as values.
Any such function $f$ describes a particular way of rating items, and corresponds to a single user.
Now, a set of observations $\{ f_1, \ldots, f_l \}$ of $F$ is referred to as \textit{functional data}.
In practice, these rating functions are not given directly, but instead, for each function $f_i$ a set of samples $\{ (x_{i;1}, f_i(x_{i;1})),\ldots,(x_{i;m_i},f_i(x_{i;m_i})) \}$ is provided.
We can put all this information together into a set with elements of the form $ (i, x_{i;j}, f_i(x_{i;j})) $,
which shows the equivalence to the rating data introduced in our problem statement (Section~\ref{sec:problem}).
So our goal of learning user embeddings essentially means that we are trying to find vector representations of functions based on lists of samples.
This corresponds to dimensionality reduction of functions, which is a sub-task of FDA \cite{wang2016functional}.

This view of the problem should make another thing clear:
User embeddings are representations of the users' rating functions.
There is no a priori justification for assuming that any other properties of the users (apart from their rating behaviors) would be incorporated into user embeddings by fitting such data.
Hence, relations between user attributes (such as gender or location) and embeddings can be useful to analyze the role of such attributes for user behavior,
but are only suitable for evaluating the embeddings if there is a known connection between rating behavior and the given user attributes.

\subsection{User similarity} \label{sec:perspective_similarity}

In Section~\ref{sec:problem} we formulated the goal that similarities of embeddings should reflect similarities of the corresponding users
(or, to be more precise, their rating behaviors, as we have just argued in the previous section). 
Before we can turn this criterion into a quantitative measure, we need to quantify similarity of users.

In the collaborative filtering literature, we find several proposed methods for computing such similarities, including
Pearson correlation, Spearman correlation, cosine vector similarity, adjusted cosine vector similarity, and mean-squared difference \cite{gong2010collaborative}.
Out of these common options we choose to estimate similarity based on the mean-squared difference of their ratings.
Our choice has two main reasons: First, mean-squared difference is most similar to the L2 distance between functions, which is a common measure used in FDA for computing distance between functions \cite{jacques2014functional}.
Second, the mean-square difference of user ratings is interpretable as expected value (assuming uniform prior over input items) of the squared difference of their ratings for the same item.

\subsection{Pair-Distance Correlation measure (PDC)} \label{sec:rank_measure}

We are not aware of any existing measure for evaluating embeddings based on function similarity.
Thus, for measuring the quality of learned embeddings, we introduce a novel performance measure,
which we call \textit{Pair-Distance Correlation} (PDC) measure. 
Two crucial elements of the presented measure are the distance metric $d_E$ on the embedding space, and the distance measure $d_U$ on the user space.

The choice for these two elements should be informed by the purpose of learning user embeddings.
For interpretability, it makes sense to consider intuitive distances on the embedding space ($d_E$) such as the Euclidean distance.
As mentioned above, we generally consider the mean-squared difference to be an appropriate choice for $d_U$,
but in certain scenarios one might want to deviate from that (e.g., mean-absolute difference if more weight should be given to small differences).
Once $d_E$ and $d_U$ are chosen, the PDC of an embedding function is computed as follows:

\begin{algorithm}[H]
\caption{Algorithm for computing Pair-Distance Correlation}
\begin{algorithmic}[1]
 \renewcommand{\algorithmicrequire}{\textbf{Input:}}
 \renewcommand{\algorithmicensure}{\textbf{Output:}}
  \REQUIRE set $\{ (x_1,u_1,R_1), \ldots, (x_n,u_n,R_n) \}$ with input items $x_i \in \mathbb{R}^n$, user identifiers $u_i \in \mathcal{U}$ and ratings $R_i \in \mathbb{R}$; embedding function $e: \mathcal{U} \rightarrow \mathbb{R}^z$; distance measures $d_E$ on $\mathbb{R}^z$ and $d_U$ on $\mathcal{U}$; threshold $t \in \mathbb{N}_{>0}$
  \ENSURE PDC score of $e$ with respect to $d_E$ and $d_U$
  \STATE set $l_E = list()$, $l_U = list()$
  \FOR {all users $u_i, u_j \in \mathcal{U}$ with at least $t$ items rated by both}
  \STATE based on all items rated by both $u_i$ and $u_j$, compute $d_U(u_i,u_j)$ and append it to $l_U$
  \STATE compute $d_E(e(u_i),e(u_j))$ and append it to $l_E$
  \ENDFOR
  \RETURN Pearson correlation coefficient between $l_E$ and $l_U$
\end{algorithmic}
\end{algorithm}

Being based on Pearson correlation, the resulting score takes values in $[-1,1]$, where higher values are preferable and $1$ is the best possible outcome.
Note that random embeddings can be expected to achieve a PDC score around $0$.
A high PDC measure (close to $1$) means, that in general if a user embedding $e(u_i)$ is more similar to $e(u_j)$ than $e(u_k)$ with respect to $d_E$ (i.e., $d_E(e(u_i),e(u_j))<d_E(e(u_i),e(u_k))$),
then the rating behavior of user $u_i$ is more similar to the behavior of $u_j$ than to that of $u_k$ (in terms of the similarity measure explained in Section~\ref{sec:perspective_similarity}, i.e., $d_U(u_i,u_j)<d_U(u_i,u_k)$). 
In other words, PDC evaluates whether an embedding function preserves distance relations.

\section{FUSION STRATEGIES} \label{sec:methods}

We would like to train a neural network that takes item information $x$ as input and user embedding $e$ as additional context signal for predicting the user's rating.
In general we distinguish between three ways for incorporating such context information into neural networks:
\begin{enumerate}
    \item \emph{neuron-level fusion}: based on context signal alter hidden states at some layer 
    \item \emph{weight-level fusion}: based on context signal alter weights of some layer
    \item combinations of the former two
\end{enumerate}

We outline three specific approaches that fall into the first two of these categories, which we will also use in our experiments later on.
Then, we will also have a closer look at Factorization Machines (FMs) \cite{rendle2010factorization}, which have shown top performance for various tasks involving user-dependent prediction \cite{rendle2012social, bayer2013factor, bayer2015fastfm}. 
The paper that introduces FMs \cite{rendle2010factorization} also explains how FMs can mimic many other popular recommendation system methods, including matrix factorization and specialized methods such as SVD++ \cite{koren2008factorization} or PITF \cite{rendle2010pairwise}.
Note that even though FM is not a neural network method per se, it can, after a small modification, be understood as weight-level fusion approach.

\subsection{Neuron-level fusion}

We focus on mask-based methods for neuron-level fusion.
In mask-based methods, a mask of the same size as the hidden state at some level is computed from the context signal and then combined with the hidden state in an element-wise manner for an update.
Two common fusion approaches are considered:
\begin{itemize}
    \item For using \textit{additive masks} (Add) on any hidden state $x$, we compute a mask of the shape of $x$ by multiplying the context vector with a weight matrix of suitable shape, and then add this mask to the original state.
    \item \textit{Multiplicative masks} (Mul) work analogously but combine mask and hidden state by element-wise multiplication (i.e., Hadamard product).
\end{itemize}

Note that additive masks are equivalent to concatenation of input $x \in \mathbb{R}^n$ and context signal $e \in \mathbb{R}^z$ one layer earlier, since it holds that
\begin{equation}
  y = W \left[\frac{x}{e}\right] = [W_1 | W_2] \left[\frac{x}{e}\right] = W_1 x + W_2 e \,,
\end{equation}
where $\left[\frac{x}{e}\right]$ stands for the concatenation of $x$ and $e$ and the $ m \times (n+z)$-matrix $W$ is split into the $m\times n$-matrix $W_1$ and the $m \times z$ matrix $W_2$.

\subsection{Weight-level fusion}

We describe \textit{tensor fusion} as one way to make weights context-dependent.
We made this choice because tensor fusion is a basic approach which we find very suitable for illustrating the general principles of weight-level fusion.
Other approaches such as the one inspired by Singular Value Decomposition (introduced for one-shot learning in \cite{bertinetto2016learning}) can typically be understood as a modification of tensor fusion.
Different from neuron-level fusion, which can normally be applied in exactly the same way in linear layers and convolutional layers,
the details of weight-level fusion depend on the type of layer.

We describe tensor fusion on a linear layer.
So let us assume that we have input $x \in \mathbb{R}^n$, context $e \in \mathbb{R}^z$ and want to map this to the output space $\mathbb{R}^m$.
The standard output of such a linear layer that ignores all context information is then
\begin{equation}
    \text{fc}(x) := b + W x \,,
\end{equation}
where $W \in \mathbb{R}^{m \times n}$ is a weight matrix and $b \in \mathbb{R}^m$ a bias term.
The basic idea of tensor fusion is to make the weight matrix $W$ dependent on the context $e$ by adding a context-dependent part to it.
More precisely, we define
$W(e) := W + eT$ where $T \in \mathbb{R}^{z \times m \times n}$ is a third-order tensor, and $eT$ is calculated as $eT = \sum_{i=1}^z e_i T_{i;\cdot;\cdot}$.
The final output of the linear tensor fusion layer is then given by
\begin{align}
  \text{tensor}(x,e) :=&\, b + W(e) x = b + (W + eT) x \\
  =&\, b + W x + e T x
\end{align}

\subsection{Factorization Machines}

Using a slightly different notation than in the original paper \cite{rendle2010factorization}, 
we can write the model equation of a FM (of degree 2 and rank $z$) as follows:
\begin{equation}
    \text{FM}(x) := b + W x + \sum_{i=1}^n \sum_{j=i+1}^n x_i V_{i;\cdot} V_{j;\cdot}^T x_j \,,
\end{equation}
where $x \in \mathbb{R}^n$, $W \in \mathbb{R}^{n}$, $V \in \mathbb{R}^{n\times z}$.
(Note that the output dimension $m$ is $1$ for FMs.)

A modified version of FM turns out to be a special case of tensor fusion, which we will now explain.
By changing the sum over $j$ to go from $1$ to $n$ (instead of $i+1$ to $n$), we get:
\begin{align} 
   \text{FM}_T(x) :=&\,  b + W x + \sum_{i=1}^n \sum_{j=1}^n x_i V_{i;\cdot} V_{j;\cdot}^T x_j \\
   =&\,  b + W x + x V V^T x
\end{align}
There are two ways how $\text{FM}_T$ can be understood as tensor fusion:
First, if we define $T := V V^T \in \mathbb{R}^{n \times n}$, we see that 
\begin{equation}
    \text{FM}_T(x) = b + W x + x T x
\end{equation}
becomes equivalent to tensor fusion that uses the same vector as input and context, and factorizes the weight tensor.
Second, we can consider $V$ as embedding matrix, so that $x$ is used as input and its embedding $x V$ as context:
\begin{equation}
    \text{FM}_T(x) = b + Wx + (xV) V^T x
\end{equation}
In this case, $V^T$ takes the role of the tensor $T$, and we have tensor fusion that shares weights with the embeddings.

We would like to point out that by interpreting the modified FM (which can still capture higher-order dynamics) in any of these two ways,
it becomes simple to see how structures similar to FM can be incorporated anywhere into a (potentially large) neural network.
In particular, these interpretations as tensor fusion explain how the method can be adapted for higher-dimensional output (where both of the two interpretations we discussed lead to slightly different adaptations).

\section{EXPERIMENT} \label{sec:experiments}

The experiment aims to evaluate the effects of embedding size and fusion strategy on the quality of user embeddings.
Apart from the baselines, we also conduct similar experiments with Factorization Machines as a benchmark. 
The experiment uses the MovieLens-100k dataset \cite{harper2016movielens}, which contains 100,000 movie ratings. 
The neural networks used in this experiment are based on linear layers.

\subsection{Task} 

We use the MovieLens-100k dataset, which consists of 100,000 movie ratings (1-5) from 943 users and of 1682 movies.
Each movie in this dataset has a unique ID and meta-information about title, year of appearance and genre(s).
None of these seem overly interesting to use as interpretation input, hence we took the movie genome information from the MovieLens-20M dataset \cite{harper2016movielens}
in order to obtain a $1128$-dimensional tag-based feature representation of the movies.\footnote{Note that for this we had to link the movie IDs between these two datasets, which we did based on movie titles and years of appearance.
We dropped the ($<$200) movies for which no corresponding movie was found.}

\begin{figure}[h]
\begin{center}
    \includegraphics[width=0.7\linewidth]{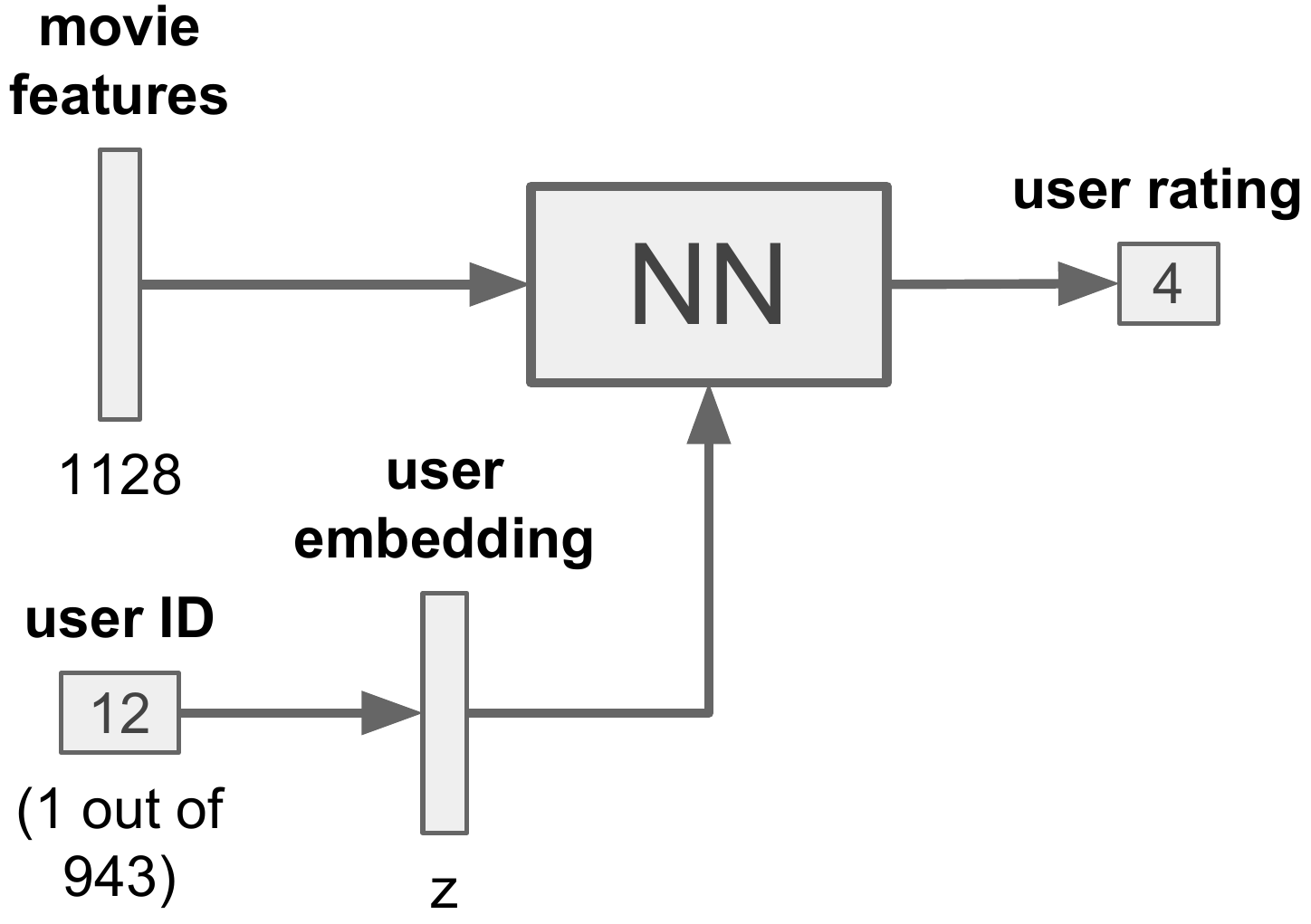}
\end{center}
\caption{Illustration of the MovieLens-100k task. A neural network is trained to take movie features as input and user ID as context signal for predicting user ratings.
The particular neural network architecture and embedding size are varied in the experiment.
Note that everything is trained end-to-end, which includes learning the embeddings.} \label{fig:movielens_task}
\end{figure}

We train various neural networks on the task of movie rating prediction, given the movie as tag feature vector and the user ID as input.
(See task illustration in Figure~\ref{fig:movielens_task}.)
It is important to recall that our main interest lies in learning meaningful embeddings. 
Embedding quality is evaluated by computing the PDC measure with respect to mean-square difference on users (estimated based on the test data)
and Euclidean distance on the embeddings (as introduced in Section~\ref{sec:rank_measure}).
Additionally, we evaluate prediction performance, using the standard recommendation system measures mean average error (MAE) and root mean squared error (RMSE).

\subsection{Architectures}
\begin{figure}[h]
\begin{center}
    \includegraphics[width=0.7\linewidth]{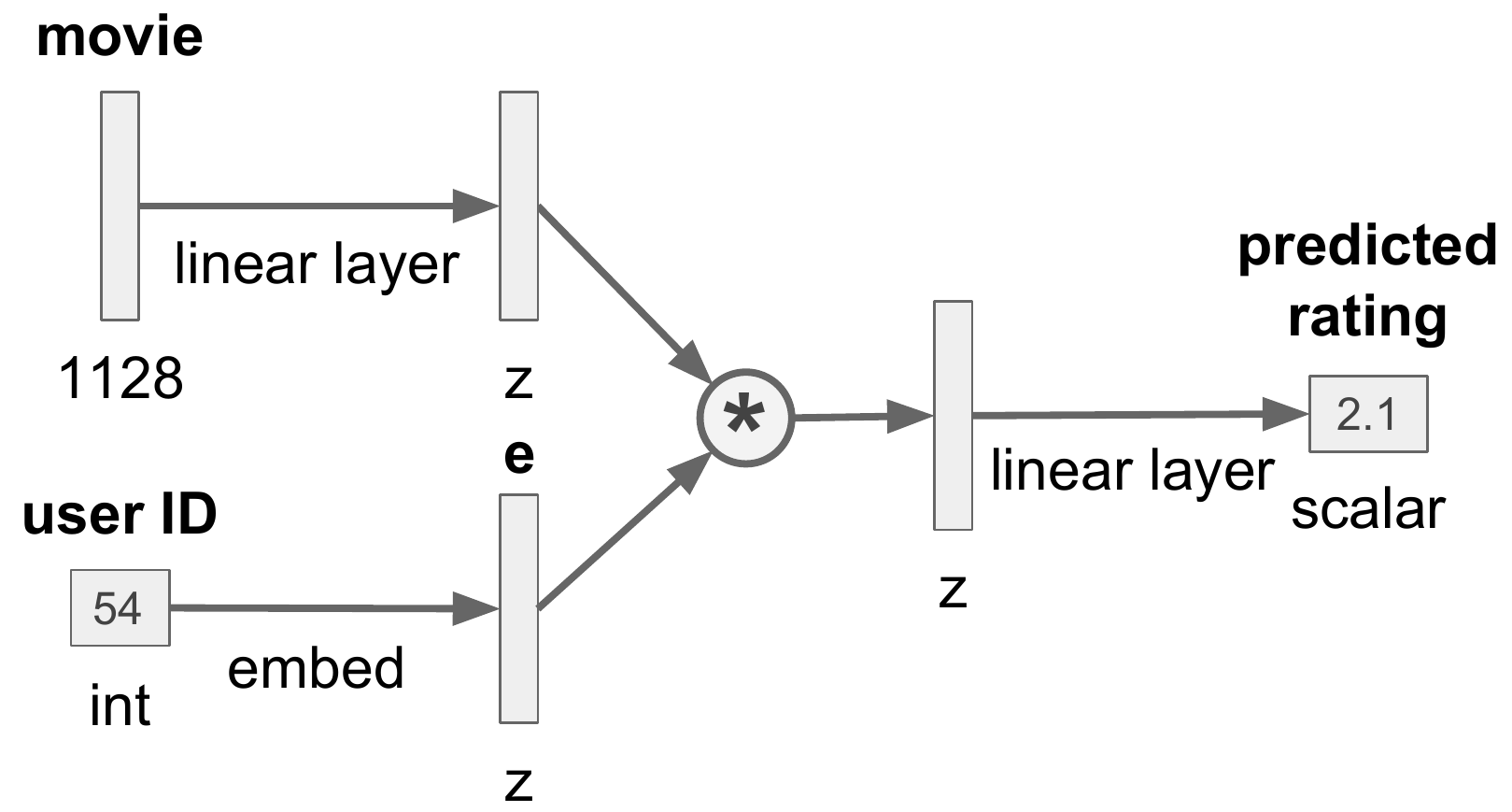}
    \includegraphics[width=0.7\linewidth]{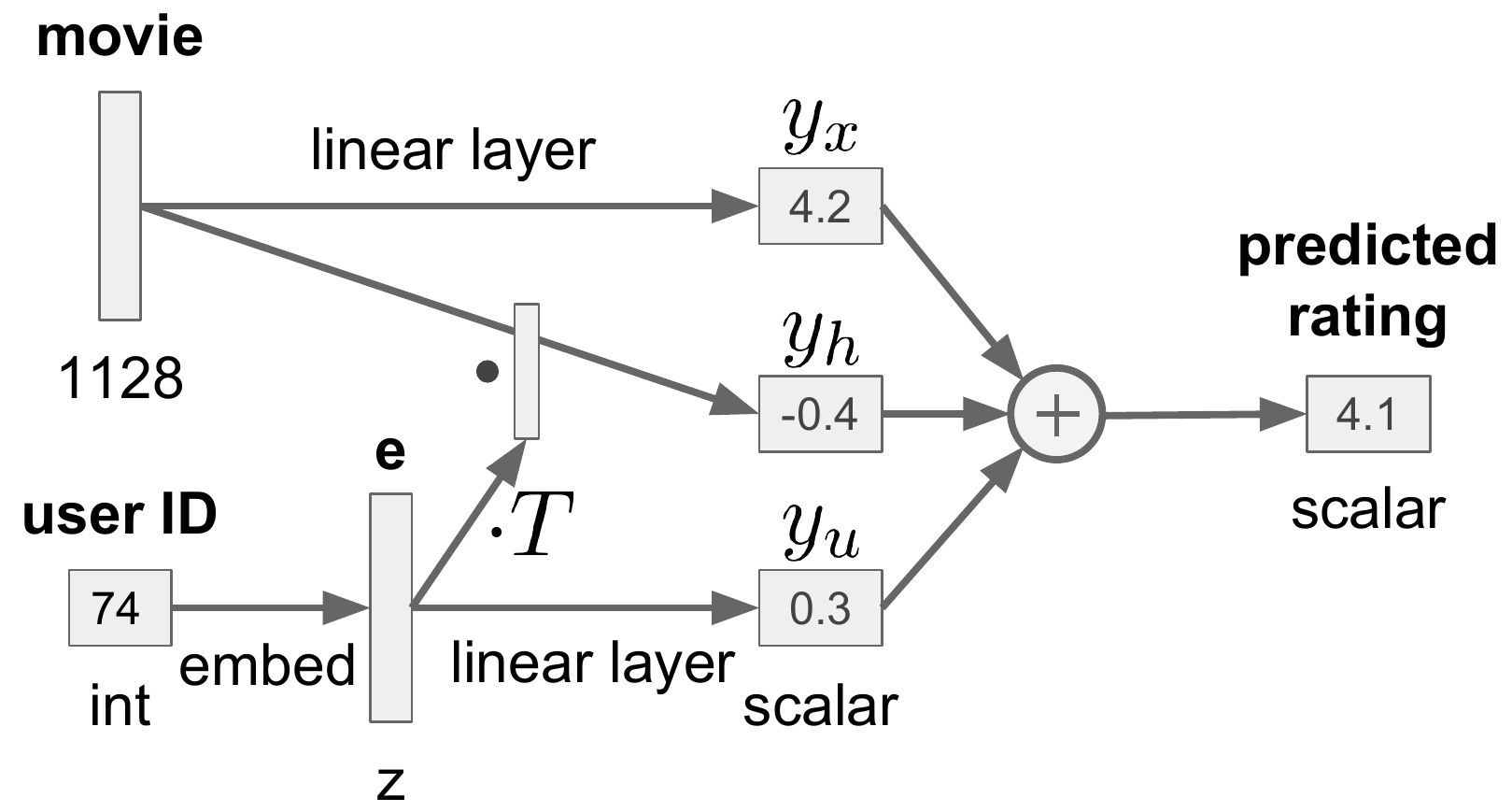}
\end{center}
\caption{Illustration of neural network architectures used for the MovieLens-100k experiment. For additive and multiplicative fusion, the basic architecture can be understood as multi-layer perceptron with one hidden layer of size $z$, where the hidden activations are modified depending on the user signal by means of element-wise addition/multiplication (taking the place of ``*''). 
As tensor fusion approach, we choose a single linear tensor fusion layer, that uses the user embeddings as context signal.} \label{fig:movielens_architectures}
\end{figure}

The neural network architectures used for this experiment are illustrated in Figure~\ref{fig:movielens_architectures}.
All of these architectures are based on one or two linear layers, and incorporate the user information by additive masks, multiplicative masks or tensor fusion, respectively (see Section~\ref{sec:methods}).
We deploy all masking mechanisms before applying the activation function.
For each of the four fusion methods, we vary the embedding size (2, 4, 8, 16, 32 and 64).

We compare these neural networks against a Factorization Machine (FM). 
For each user, an embedding in case of FM is obtained by appending the user bias (as learned by $W$) to the row of the weight matrix $V$ which corresponds to the user.

\subsection{Results}
\begin{figure*}[h]
\begin{center}
    \includegraphics[width=\linewidth]{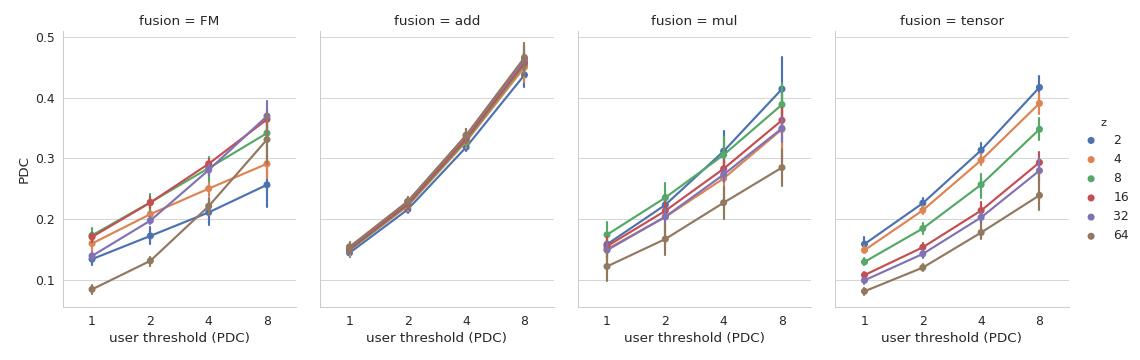}
\end{center}
\caption{PDC scores for embeddings of different fusion strategies for various embedding sizes. The threshold that is varied along the x-axis is a parameter of the PDC measure: For computing the PDC measure, only user pairs with this number of common movies are considered. Mean values and standard deviation are calculated based on training and evaluation of 5 folds. Corresponding scores of the linear baseline are 0.12, 0.19, 0.29, 0.42 (thresholds 1, 2, 4, and 8, respectively).} \label{fig:embedding_behavior}
\end{figure*}

Results can be found in Table~\ref{tab:movielens_results}.
As two baselines, we include a model that outputs the average rating of the given user and ignores the movie features (user-bias),
and a model that adjusts this score based on average user ratings by adding a learned linear combination of the movie features (linear).
For the baselines, the (scalar) user-dependent biases were used as embeddings.
All reported results are averages of 5-fold cross validation, using the official dataset split.
In Figure~\ref{fig:embedding_behavior}, we additionally show how embedding scores vary across fusion strategies and embedding sizes, depending on the threshold of common items that is used for computing the PDC score.
We optimized hyper-parameters (learning rate, number of epochs, and in case of FM also the regularization terms) 
based on a different split.
We implemented all baselines and neural network models in TensorFlow \cite{abadi2016tensorflow}, and used Sacred \cite{greff2017sacred} for managing our experiments.
FMs were trained and evaluated in a separate script, using Bayer's fastFM implementation \cite{bayer2015fastfm}.
\begin{table}[t]
\caption{Prediction (MAE, RMSE) and embedding scores (PDC) of the presented fusion strategies and baselines. PDC scores are computed with respect to mean-square difference between users and Euclidean distance on embeddings. For calculating the PDC scores, only user pairs with at least 4 common ratings were considered (threshold 4). The +1 in the z column represents the bias term.}
\begin{center}
\begin{tabular}{lc|cc|c|r}
\multicolumn{2}{c|}{\bf Approach}  &\multicolumn{2}{c|}{\bf Prediction} & {\bf PDC} 
&  \\
{\bf Fusion} & {\bf z} & {\bf MAE} & {\bf RMSE} & {\bf (threshold 4)} & {\bf Params} \\
\hline 
user-bias & 0+1          & 0.87 & 1.06 & 0.26 & 946 \\
\hline
linear & 0+1          & 0.76 & 0.95 & 0.29 & 3015 \\
\hline
        & 2+1          & 0.71 & 0.91 & 0.26 & 6214 \\
        & 4+1          & 0.71 & 0.91 & 0.29 & 10356 \\
FM  & 8+1          & 0.72 & 0.92 & 0.31 & 18640 \\
        & 16+1          & 0.75 & 0.96 & 0.29 & 35208 \\
        & 32+1        & 0.80 & 1.03 & 0.28 & 68344 \\
        & 64+1        & 0.83 & 1.07 & 0.24 & 134616 \\
\hline 
       & 2          & 0.74 & 0.94 & 0.32 & 4147 \\
       & 4          & 0.74 & 0.93 & 0.33 & 8293 \\
add    & 8          & 0.73 & 0.93 & 0.33 & 16585 \\
       & 16          & 0.73 & 0.93 & 0.33 & 33169 \\
       & 32          & 0.74 & 0.93 & {\bf 0.34} & 66337 \\
       & 64          & 0.74 & 0.94 & {\bf 0.34} & 132673 \\
\hline 
       & 2          & 0.73 & 0.92 & 0.31 & 4147 \\
       & 4          & 0.71 & 0.91 & 0.27 & 8293 \\
mul    & 8          & \textbf{0.70} & \textbf{0.90} & 0.31 & 16585 \\
       & 16          & \textbf{0.70} & \textbf{0.90} & 0.28 & 33169 \\
       & 32          & \textbf{0.70} & \textbf{0.90} & 0.27 & 66337 \\
       & 64          & 0.71 & \textbf{0.90} & 0.23 & 132673 \\
\hline 
 & 2          & 0.73 & 0.92 & 0.31 & 5273 \\
 & 4          & 0.72 & 0.91 & 0.30 & 9417 \\
tensor & 8          & 0.71 & 0.91 & 0.26 & 17705 \\
 & 16          & 0.71 & \textbf{0.90} & 0.21 & 34281 \\
 & 32          & 0.71 & \textbf{0.90} & 0.20 & 67433 \\
 & 64          & 0.71 & 0.91 & 0.18 & 133737 \\
\end{tabular}
\end{center} \label{tab:movielens_results}
\end{table}

\subsection{Analysis}
In Table~\ref{tab:movielens_results} we see that fusion with multiplicative masks of a moderate embedding size (around 16) works best for prediction.
Tensor fusion with similar embedding sizes (16, 32) achieves comparable prediction results.
Additive masks yield comparatively poor prediction performances, but achieve higher embedding scores.
Both of these parts are largely independent of the chosen embedding size in case of addition, while we can observe a slight trend of increasing embedding quality with growing embedding size.
For tensor fusion on the other hand, 
embedding quality is overall lower, and heavily depends on the chosen embedding size, while lower dimensions yield higher quality.
Interestingly, tensor fusion favors high embedding sizes for prediction,
which means that these two types of performances are found to be anti-correlated.
Multiplicative fusion shows similar trends to those of tensor fusion, so it also prefers low embedding size for learning embeddings, but this effect is somewhat less pronounced.
Factorization Machines are highly competitive for prediction when using small embedding sizes, but at around $z=16$ start overfitting quite heavily.
Embedding scores are comparable to those of the multiplicative fusion model.
Again, optimal embedding size is different for prediction as compared to embedding quality (8 vs 2 or 4). 
In general it is surprising that there seems to be no clear relation between prediction performance and embedding quality (compare, e.g., addition with $z=32$, multiplication with $z=4$ and tensor fusion with $z=64$).

Results for PDC score in Table~\ref{tab:movielens_results} are all based on a user threshold of $4$, i.e., only pairs of users with at least $4$ commonly-rated movies were considered for the calculation.
In Figure~\ref{fig:embedding_behavior} we can see how embedding scores change if we vary this user threshold.
Note that for lower thresholds we have many more user pairs to consider but also much more noise in the data,
which explains why scores are generally much lower for lower user thresholds.
Most of the effects we discussed above are insensitive to this user threshold we chose for computing the PDC score.
One exception to this insensitivity is the observation that for low user thresholds, addition is on par with some tensor fusion models, and even slightly outperformed by certain variants of FM and multiplicative fusion ($z=8$).
Together with the corresponding prediction performances, this suggests that additive fusion generally focuses on less complex user-dependent effects (as compared to FM and multiplicative fusion).
Figure~\ref{fig:embedding_behavior} also reveals a high standard deviation of the embedding quality for multiplicative fusion.
This is generally an undesired property but could indicate that multiplicative fusion might benefit from additional regularization techniques.

Overall, in this experiment additive fusion appears to be a robust choice for learning high quality embeddings, as long as prediction performance is not important.

\subsection{Input sensitivities and user clustering}
The model-based approach we adopted for fitting user data while simultaneously learning to represent the users as embedding vectors gives us one other interesting option,
which we have not yet mentioned:
For any user embedding, the trained model describes the associated rating behavior, which we can analyze further.
In particular, for a given user, we can compute input sensitivities (partial derivatives of rating score with respect to input features)
in order to find out, which features of a movie make the user more likely to rate the movie higher, and which features the user does not like.
Of course, we should not assume that any of our models perfectly fits the true rating behavior of any user, especially in terms of more complex properties such as input sensitivities.
Still, if the models achieve reasonable prediction performance, there is good reason to believe that at least some properties are captured correctly.
And choosing a model with a simple structure furthermore reduces the chance of ending up with complex statistical artifacts.

Among our neural network architectures, the structure of the tensor fusion model is particularly simple, which even allows for direct interpretation of the learned weights (see Figure~\ref{fig:movielens_architectures}).
The model has a user-independent rating prediction $y_x$ based on the movie features alone.
To this baseline prediction, two numbers are added that both depend on the user.
The first number is a general user bias $y_u$, computed from the user embedding. For the second number $y_h$, the user embedding is mapped to a vector which then serves as weights to compute another linear combination of the movie features.
This second weight vector directly describes user-dependent changes in input sensitivity, since there is no non-linearity in the model.
Hence, for any user embedding we obtain a corresponding bias term and changes in input sensitivities without having to put any item data through the model.

This looks very different for multiplicative or additive fusion, where user-dependent effects on input sensitivities can vary across items.
Input sensitivities (or relevancies) can still be analyzed in this case by using heatmapping techniques (e.g., \cite{MONTAVON20181,SamITU18}),
but here the risk of observing statistical artifacts becomes higher (since the possibility that input sensitivities can differ across items drastically increases the number of effects to analyze).

To at least get an intuition of how helpful such input sensitivities might be in practice, we run another experiment as preliminary analysis.
In this experiment, we select the tensor fusion model with embedding size 4 (of best performing training fold) and have a closer look at what the model has learned.
To this end, we run k-means clustering with 20 clusters on the learned user embeddings. For $3$ random clusters, we pick the centroid embedding and read the associated biases and changes in input sensitivities from the model.
We also include the highest- and lowest-ranked movies based on these values.
The results can be found in Table~\ref{tab:user_clusters}.
\begin{table*}[t] 
\caption{Biases, favorite and least liked movie features and movies associated with centroids of 3 random user clusters. Clustering was done on user embeddings learned by the tensor fusion model with embedding size 4.
The biases and scores of movie features were read from the same model, which was also used for ranking the movies.
The table contains the 5 highest/lowest ranked features and 3 highest/lowest ranked movies, respectively.}
\begin{center}
\begin{tabular}{c|c|p{.18\textwidth}|p{.20\textwidth}|p{.18\textwidth}|p{.20\textwidth}}
{\bf Cluster} & {\bf General} & \multicolumn{2}{c|}{\bf Highest ranked} & \multicolumn{2}{c}{\bf Lowest ranked} \\
{\bf No.} & {\bf bias} & {\bf Movie features} & {\bf Movies} & {\bf Movie features} & {\bf Movies} \\
\hline
1 & -0.019 & italy, unlikeable characters, road trip, character study, parody & 
             Pulp Fiction (1994), A Clockwork Orange (1971), The Big Lebowski (1998) &
            women, nudity, history, marx brothers, great & 
             Between the Folds (2008), Duma (2005), McFarland USA (2015) \\
\hline
2 & -0.033 & visuals, dark humor, classic, non-linear, sarcasm & 
             Pulp Fiction (1994), Reservoir Dogs (1992), Taxi Driver (1976) &
            predictable, childhood, betrayal, cheating, bad acting & 
             You've Got Mail (1998), Ghost (1990), Runaway Bride (1999) \\
\hline
3 & 0.017 & intimate, visually appealing, costume drama, women, whimsical & 
             Cries and Whispers (1972), Last Life in the Universe (2003), Submarino (2010) &
            chase, 70mm, snakes, teleportation, chris tucker & 
             Independence Day (1996), Transformers (2007), Men in Black (1997) \\
\end{tabular}
\end{center} \label{tab:user_clusters}
\end{table*}
The constellation of features and movies in these results seems coherent and suggests that this is a promising direction for further investigation.

\section{CONCLUSION} \label{conclusion}

In this paper, we introduced the PDC measure for evaluating user embeddings based on similarities of their rating behavior.
This novel measure formalizes the intuitive requirement that similar users should be mapped to similar vectors.

We conducted an experiment on movie rating data, where we compared additive, multiplicative, and tensor fusion in neural networks that learn to fit this data while forming vector representations of all the users.
In our experiment we found that the fusion strategy has a significant effect on prediction as well as the quality of the learned embeddings.
The effect of embedding size on prediction performance and embedding quality seems to largely depend on the chosen fusion strategy.
Additive conditioning was mostly unaffected by changes in embedding size and other methods generally favored small embedding sizes for high embedding quality. 
Surprisingly, good prediction performance does not necessarily reflect the quality of the learned embeddings.
In case of tensor fusion, we even observed these two aspects to be anti-correlated.

This is an important finding since one tends to select models based on their prediction ability,
but apparently it is not at all clear how well this measure correlates with other aspects of interest, such as ``meaningfulness'' of embeddings or learned input sensitivities.
In our opinion, this finding suggests that much more work is necessary to better understand the internal dynamics of neural networks,
especially when fusion of different information is involved and the models are to be used for data analysis. 

Finally, it shall be mentioned that,
although we formulated the problem in terms of user ratings,  the same modeling can directly be applied to other data such as dialogues.
In fact, our chosen approach for learning user embeddings fits the theoretical framework of interpretation analysis proposed by  \cite{blandfort2018overview},
and can be seen as a case of model-based interpretation analysis. 

\section*{ACKNOWLEDGMENT}

This work was supported by the BMBF project DeFuseNN (Grant 01IW17002) and the NVIDIA AI Lab (NVAIL) program.
Furthermore, the first author received financial support from the Center for Cognitive Science, Kaiserslautern, Germany.


\bibliographystyle{ieeetr}
\bibliography{main}

\end{document}